\documentclass[lettersize,journal]{IEEEtran}
\usepackage{amsmath,amsfonts}
\usepackage{algorithmic}
\usepackage[american]{babel}
\usepackage{array}
\usepackage[caption=false,font=normalsize,labelfont=sf,textfont=sf]{subfig}
\usepackage{textcomp}
\usepackage{stfloats}
\usepackage{url}
\usepackage{verbatim}
\usepackage{xcolor}
\usepackage{booktabs}
\usepackage{multirow, makecell, xcolor, colortbl}
\usepackage{changepage}
\usepackage{adjustbox}
\usepackage{array}
\usepackage{hyperref}
\hypersetup{
    colorlinks=true,      
    linkcolor=blue,       
    citecolor=blue,       
    urlcolor=black,        
    filecolor=blue,       
    linkbordercolor={0 0 0}, 
    pdfborder={0 0 0}     
}
\usepackage{orcidlink}

\usepackage{graphicx}
\def\BibTeX{{\rm B\kern-.05em{\sc i\kern-.025em b}\kern-.08em
    T\kern-.1667em\lower.7ex\hbox{E}\kern-.125emX}}
\usepackage{balance}
\begin{document}
\title{Enhancing Object Detection with Privileged Information: A Model-Agnostic Teacher–Student Approach}
\author{%
Matthias~Bartolo~\orcidlink{0009-0006-1353-4556}, 
\and
Dylan~Seychell~\orcidlink{0000-0002-2377-9833},~\IEEEmembership{Senior Member,~IEEE}, %
\and
Gabriel~Hili~\orcidlink{0009-0003-5996-4189}, %

\and
Matthew~Montebello~\orcidlink{0000-0002-6177-7548},~\IEEEmembership{Senior Member,~IEEE}, %
\and
Carl~James~Debono~\orcidlink{0000-0003-2659-8752},~\IEEEmembership{Senior Member,~IEEE}, %

\and
Saviour~Formosa~\orcidlink{0000-0002-4449-1791}, %
\and
and Konstantinos~Makantasis~\orcidlink{0000-0002-0889-2766},~\IEEEmembership{Member,~IEEE}

\thanks{%
\textit{(Corresponding author: Matthias Bartolo.)}\\
Matthias Bartolo, Dylan Seychell, Gabriel Hili, Matthew Montebello, and Konstantinos Makantasis are with the Department of Artificial Intelligence, Faculty of Information and Communications Technology, University of Malta, MSD 2080, Malta (e-mail: matthias.bartolo@um.edu.mt; dylan.seychell@um.edu.mt; gabriel.hili@um.edu.mt; matthew.montebello@um.edu.mt; konstantinos.makantasis@um.edu.mt).
Carl James Debono is with the Department of Communications and Computer Engineering, Faculty of Information and Communications Technology, University of Malta, MSD 2080, Malta (e-mail: carl.debono@um.edu.mt).
Saviour Formosa is with the Department of Criminology, Faculty of Social Wellbeing, University of Malta, MSD 2080, Malta (e-mail: saviour.formosa@um.edu.mt).
}%
\thanks{
This work is part of the project \textit{‘Aerial Waste Identification and Geolocation System’ (AWIGS)}, financed by Xjenza Malta, through the FUSION: R\&I Technology Development Programme Lite.}
}

\markboth{Submitted to IEEE TRANSACTIONS ON IMAGE PROCESSING,~Vol.~N/A, No.~N/A, January~2026}%
{How to Use the IEEEtran \LaTeX \ Templates}

\maketitle

\begin{abstract}
This paper investigates the integration of the Learning Using Privileged Information (LUPI) paradigm in object detection to exploit fine-grained, descriptive information available during training but not at inference. We introduce a general, model-agnostic methodology for injecting privileged information—such as bounding box masks, saliency maps, and depth cues—into deep learning-based object detectors through a teacher–student architecture. Experiments are conducted across five state-of-the-art object detection models and multiple public benchmarks, including UAV-based litter detection datasets and Pascal VOC 2012, to assess the impact on accuracy, generalization, and computational efficiency. Our results demonstrate that LUPI-trained students consistently outperform their baseline counterparts, achieving significant boosts in detection accuracy with no increase in inference complexity or model size. Performance improvements are especially marked for medium and large objects, while ablation studies reveal that intermediate weighting of teacher guidance optimally balances learning from privileged and standard inputs. The findings affirm that the LUPI framework provides an effective and practical strategy for advancing object detection systems in both resource-constrained and real-world settings.
\end{abstract}


\section{Introduction}
\IEEEPARstart{A}{dvancements} in computing hardware, particularly GPUs, have enabled the rapid adoption of artificial intelligence and automation technologies. Within this landscape, object detection has emerged as a cornerstone problem, driving applications in areas such as autonomous systems, environmental monitoring, and robotics. Over the past decade, models such as YOLO \cite{yolo}, Faster R-CNN \cite{fasterrcnn}, and RetinaNet \cite{retinanet} have delivered fast and accurate detection capabilities, making object detection a widely deployable technology. Despite this progress, achieving consistently high detection accuracy remains a challenge. Many state-of-the-art detectors rely on increasingly complex architectures \cite{detr, rt-detr}, which still need to be fine-tuned for specific domain use cases using large annotated datasets \cite{od_survey_problems}, both of which introduce significant practical constraints. Deep models often require extensive training time and computational resources, while large-scale datasets demand costly and labour-intensive annotation to improve detection accuracy \cite{od_survey_problems, od_problem}.

\begin{figure}[!t]
    \centering
    \includegraphics[width=\columnwidth]{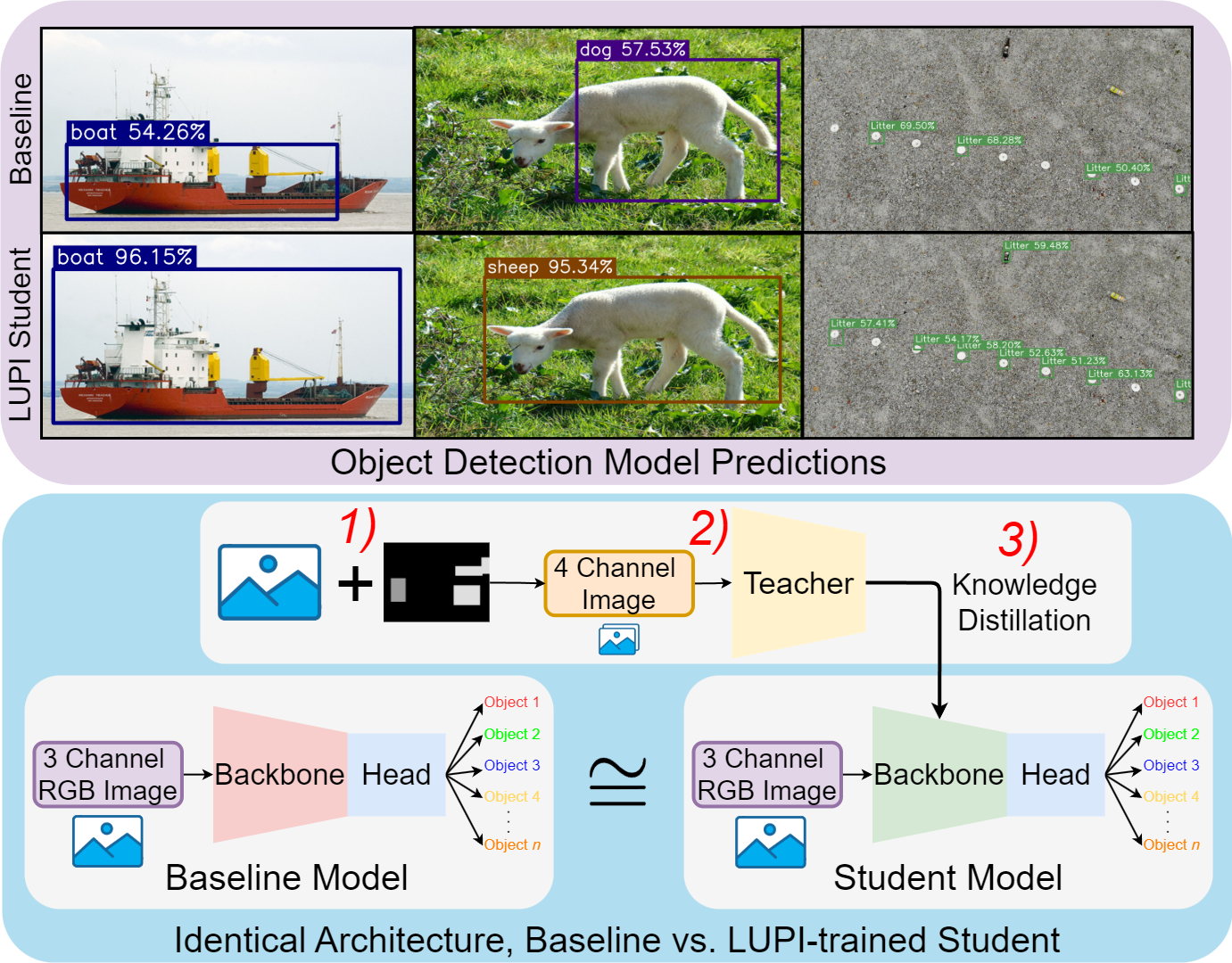}
    \caption{Visual comparison of baseline object detection predictions and those of a LUPI-trained student model, showing improved accuracy while keeping the same architecture. The figure also illustrates the LUPI training pipeline, including privileged information, teacher models, and knowledge distillation, with these boosts arising solely from the bolstered learning process.}
    \label{fig:front1}
\end{figure}

However, annotated images contain highly rich information that current state-of-the-art object detection models do not fully exploit. In this study, we test the hypothesis that highly descriptive, fine-grained information can be automatically constructed and leveraged during training to improve object detector performance. Building on our preliminary results \cite{Euvip_Paper}, we adopt the Learning Under Privileged Information (LUPI) paradigm \cite{vapnik2009new, pechyony2010theory, vapnik2015learning} and tailor its components for effective use within object detection.

The LUPI paradigm addresses problems where information asymmetry exists between training and testing: supplementary information is available during training but not during inference. By leveraging highly informative data streams that are inaccessible at test time, LUPI significantly reduces the requirement for large annotated datasets without sacrificing model accuracy. Privileged information can take many forms, including depth cues, saliency maps, high-resolution imagery, or domain-specific annotations \cite{lupi, lupi_classification, lupiv3}. By incorporating such signals, models learn richer feature representations during training, improving generalization and accelerating convergence while maintaining unchanged inference requirements (see Figure \ref{fig:front1}).

Our work is novel is several ways. First, we propose and develop a general methodology for injecting privileged information into any deep learning-based object detector. The proposed methodology is model-agnostic and not restricted by architectural choices. Second, we investigate the impact of our methodology across five open-source state-of-the-art pretrained object detection models using multiple UAV-based litter detection datasets and the Pascal VOC benchmark. Third, we build upon and significantly extend our earlier work \cite{Euvip_Paper} by analyzing performance across object scales, standard COCO metrics, and different forms of privileged information---including depth, saliency, and their combinations---while also examining practical factors such as inference time and model size. Finally, through extensive experimental validation, we demonstrate the importance of privileged information for boosting model performance and provide deeper insights into the viability of our LUPI-based approach in generic object detection, assessing both the scientific and practical implications of this paradigm.

\section{Related Work}

Object detection is a complex problem that involves both classification and localization \cite{od_problem}. The field has a rich history, evolving from early works using traditional feature matching \cite{bay2006surf, rublee2011orb} and machine learning techniques \cite{viola2001rapid, bhatt2023state} to the incorporation of deep learning methods \cite{deep_learning_review}, which currently provide state-of-the-art performance. This section reviews related studies on deep learning-based object detection and LUPI for computer vision applications.

\subsection{Deep Learning for Object Detection}
Object detection is a supervised learning task that encompasses several key challenges. These include detecting objects against complex backgrounds and interferences, accounting for scale variability, handling occlusion, mitigating class imbalance and dataset bias, and detecting small objects \cite{od_survey_problems, four_pillars_od}. All of these challenges require robust learning algorithms. Current state-of-the-art object detection models leverage various deep learning architectures that produce outputs in the form of bounding boxes accompanied by categorical labels. These networks are generally categorized into four groups: one-stage, two-stage, transformer-based, and other deep learning approaches.

One-stage detectors solve the localization and classification problems using a single network. Popular examples include YOLO (You Only Look Once) \cite{yolo} and SSD (Single Shot MultiBox Detector) \cite{ssd}. Two-stage detectors use separate networks to perform localization and classification, with examples such as Faster R-CNN \cite{fasterrcnn} and Mask R-CNN \cite{maskrcnn}. Transformer-based detectors, such as DETR \cite{detr} and RT-DETR \cite{rt-detr}, leverage self-attention mechanisms for object detection. Other deep learning approaches, like CenterNet \cite{centernet}, or SAHI \cite{sahi_detection}  incorporate reinforcement learning techniques. 

These diverse approaches highlight the range of strategies developed to tackle the challenges of object detection. One-stage detectors prioritize speed and efficiency, while two-stage detectors excel in accuracy for complex scenes. Transformer-based models offer flexibility in modeling object relationships, demonstrating that different architectural paradigms address different aspects of the detection problem.

\subsection{Learning Using Privileged Information in Computer Vision}

The use of LUPI in computer vision remains relatively underexplored, particularly within object detection. Our earlier work \cite{Euvip_Paper} represents one of the first contributions in this area. Early efforts in the literature focused on object localization tasks. Feyereisl et al. (2014) \cite{lupi_od_extra1} used segmentation masks and SURF features as privileged information for object localization using the Structural SVM+ algorithm on the Caltech-UCSD Birds dataset \cite{birds_dataset}. Improvements were marginal, and deep learning models were not yet widely adopted at the time. Similarly, Sun et al. (2018) \cite{lupi_od_extra2} examined object localization with privileged information on the same dataset, achieving limited improvements.

LUPI has seen broader exploration in image classification. Sharmanska et al. \cite{learning2rank, learning2rank2} investigated semantic attributes, bounding boxes, textual descriptions, and annotator rationale as privileged information, showing measurable improvements within the SVM+ framework. Wang et al. \cite{lupi_classification} incorporated privileged data, such as high-resolution images and tags, into multi-label classification, demonstrating that leveraging such information enhances performance. Makantasis et al. \cite{makantasis2021privileged, makantasis2023lab} used audio and physiological features as privileged information for developing vision-based models of affect in an attempt to bridge the gap between in-vitro and in-vivo affect modeling tasks.

LUPI is closely related to knowledge distillation \cite{lopez2015unifying}, a technique that has gained increasing popularity in recent years, particularly within computer vision \cite{distillation1}. In this framework, a high-capacity network (the teacher) transfers information to a smaller network (the student), enabling the student to learn richer and more informative representations \cite{distillation2}. Hinton’s concept of \textit{generalized distillation} \cite{hinton_distillation} formalizes this process. Computer vision applications commonly employ methods such as feature and logit matching between student and teacher networks, while localization distillation specifically targets spatially informative regions, helping student models focus on the most relevant areas and thereby improving detection performance \cite{distillation1, distillation2, hinton_distillation}.

However, knowledge distillation and LUPI differ fundamentally in their objectives and information requirements. The main objective of knowledge distillation is to build a compact student that performs on par with a much larger teacher model, where both models use identical input information. In contrast, LUPI \cite{vapnik2009new, pechyony2010theory, vapnik2015learning} aims not to compress a large network but to transfer knowledge from a teacher model trained using highly informative privileged information to a student model that makes predictions in the absence of that privileged information. Thus, while knowledge distillation addresses model compression, LUPI addresses information asymmetry between training and inference.

\section{Methodology}

\begin{figure*}[!t]
\centering
\includegraphics[width=1\textwidth]{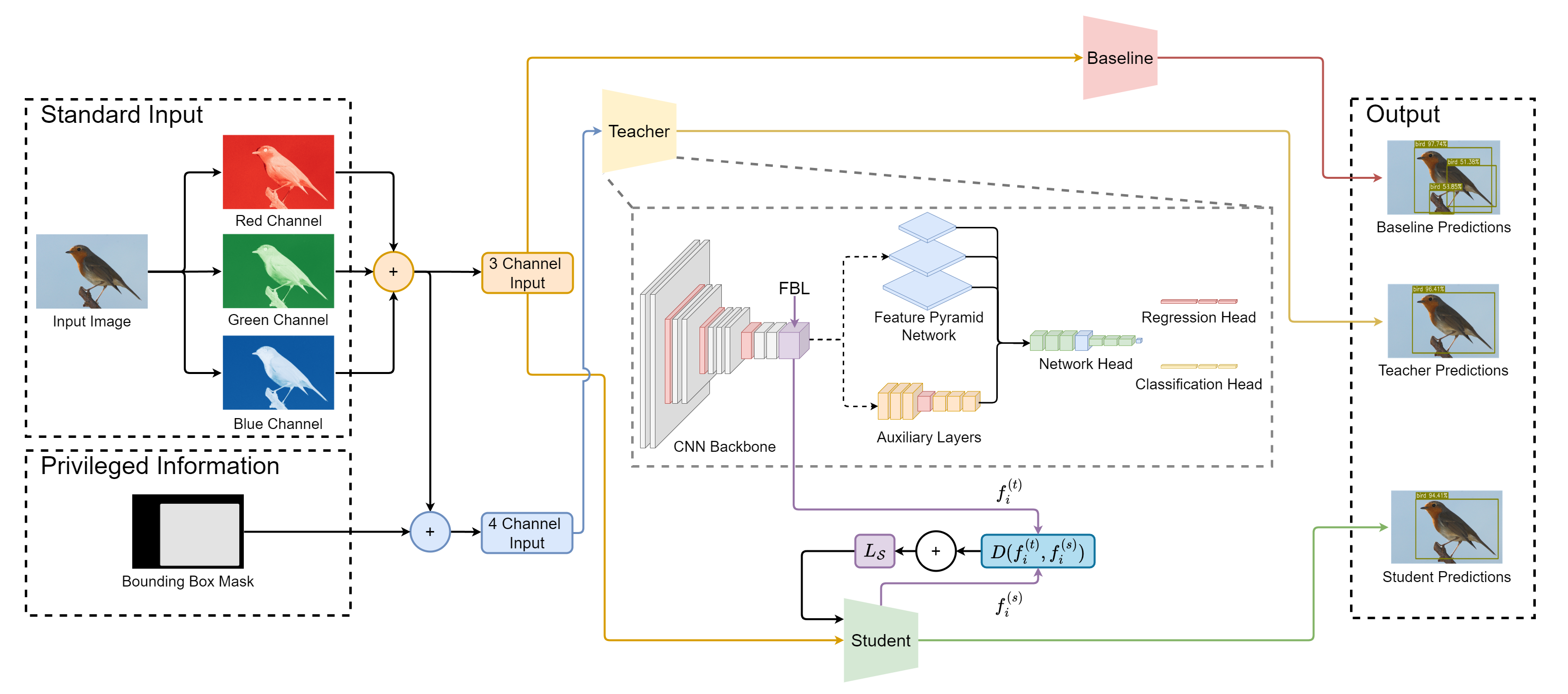}
\caption{Detailed architecture of the training setup. The teacher network receives both RGB images and privileged input channels, producing richer intermediate representations. The student network only processes RGB images, but is trained with additional supervision through knowledge distillation from the teacher. A baseline RGB-only model is included for comparison. The student demonstrates refined predictions relative to the baseline.}
\label{fig:lupi_architecture_final}
\end{figure*}

To test our hypothesis that highly descriptive, fine-grained information can be automatically constructed and leveraged during training to improve object detector performance, we follow the work in \cite{vapnik2015learning} and employ a teacher-student framework. This section formalizes the problem and describes the proposed approach.

\subsection{Problem Formulation}

We consider a supervised object detection setting where each training sample consists of a standard input image \(x \in X\), additional privileged information \(x^* \in X^*\) available only during training, and the corresponding ground-truth label \(y \in Y\), which includes a bounding box $b$ and a class label $l$ per depicted object. The training dataset is therefore a set of triplets
\begin{equation}
\mathcal{D}_{\text{train}} = \{(x_i, x_i^*, y_i)\}_{i=1}^{N},
\end{equation}
with the ground-truth label $y_i$ for image $i$ to be the set
\begin{equation}
\label{eq:labels}
y_i = \{(b_j, l_j)\}_{j=1}^{M}.
\end{equation}
In (\ref{eq:labels}), $M$ stands for the number of depicted objects in the image. Our objective is to estimate a function 
\begin{equation}
\label{eq:student}
 f_w:X\rightarrow Y    
\end{equation}
parameterized by $w$ such that 
\begin{equation}
\label{eq:weights}
    w:=w(X,X^*,Y).
\end{equation}
In our case, the function $f_w$ is implemented by a neural network and the parameters $w$ correspond to the network's weights. Equations (\ref{eq:student}) and (\ref{eq:weights}) demonstrate that the network makes predictions using only $X$, while its parameters are estimated using not only $X$ and $Y$, but also the additional privileged information $X^*$.

\subsection{Proposed Approach}

We leverage privileged information by adopting the teacher–student paradigm. The teacher network \(f_{\text{teacher}}: X \cup X^* \rightarrow Y\) has access to both standard and privileged inputs, allowing it to learn richer and more informative intermediate representations. In contrast, the student network \(f_{\text{student}}: X \rightarrow Y\) observes only the standard inputs and has no direct access to the privileged information. During training, however, the student is encouraged to replicate the teacher’s latent representations at an intermediate layer \(l\), hereby benefiting indirectly from the additional privileged context.

Both \(f_{\text{teacher}}\) and \(f_{\text{student}}\) are implemented as neural networks composed of \(L\) layers:
\begin{equation}
f_{\text{teacher}} := f_1^{(t)} \circ f_2^{(t)} \circ \cdots \circ f_l^{(t)} \circ \cdots \circ f_L^{(t)},
\end{equation}
\begin{equation}
f_{\text{student}} := f_1^{(s)} \circ f_2^{(s)} \circ \cdots \circ f_l^{(s)} \circ \cdots \circ f_L^{(s)}.
\end{equation}
Here, “\(\circ\)” denotes function composition, and \(f_i^{(t)}\), \(f_i^{(s)}\) represent the \(i\)-th layer of the teacher and student networks, respectively. The \(l\)-th layer of both networks is constrained to have the same number of hidden neurons, enabling direct comparison between their latent representations.

For each triplet $(x_i, x_i^*, y_i) \in \mathcal{D}_{\text{train}}$, the student is trained to align its latent features \(f_l^{(s)}(x_i)\) with the corresponding teacher features \(f_l^{(t)}(x_i, x_i^*)\). This alignment forms the basis of the knowledge transfer process, allowing the student to approximate the teacher’s richer intermediate representations while relying solely on standard inputs.

The student is optimised with a combined loss function that balances standard detection supervision with knowledge transfer from the teacher:
\begin{equation}
\label{eq:lupi-loss}
L_{\mathcal{S}} = (1-\alpha) \cdot L_{\text{det}} + \alpha \cdot D(f_l^{(t)}, f_l^{(s)}),
\end{equation}
where \(L_{\text{det}}\) is the standard detection loss and \(D(\cdot)\) measures the cosine distance between teacher and student feature vectors:
\begin{equation}
\label{eq:cosine-distance}
D(f_l^{(t)}, f_l^{(s)}) = 1 - \frac{f_l^{(t)} \cdot f_l^{(s)}}{\left\|f_l^{(t)}\right\| \left\|f_l^{(s)}\right\|}
\end{equation}
In (\ref{eq:lupi-loss}), \(\alpha \in [0,1]\) controls the relative weight between supervision from ground-truth labels and guidance from the teacher. The success of the teacher-student framework is closely tied to the type and quality of privileged information used during training. We next describe how this information is constructed and integrated to bolster the learning process.

\subsection{Privileged Information for Object Detection}
\label{sec:4_privileged_information}

\begin{figure*}[!t]
\centering
\includegraphics[width=1\textwidth]{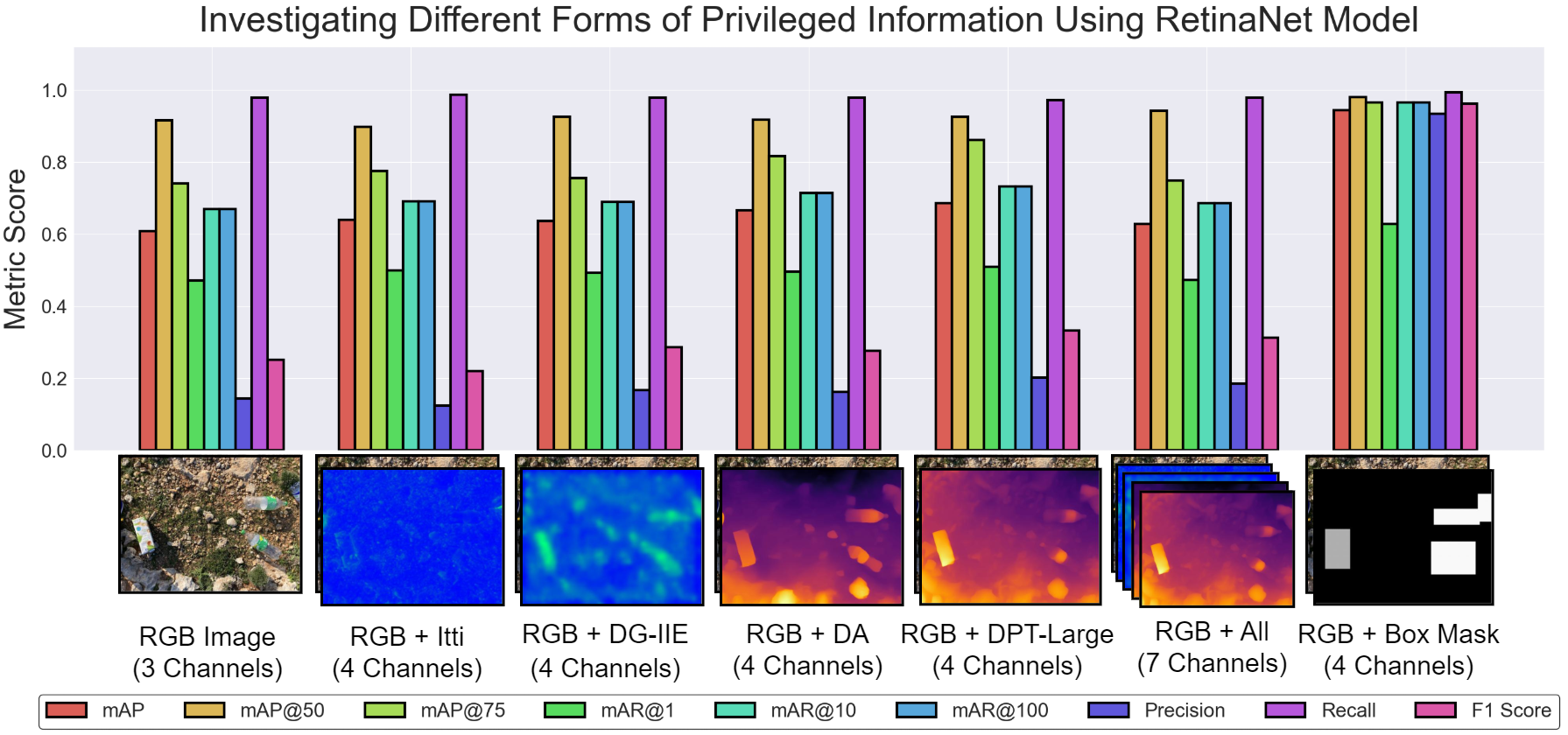}
\caption{Investigation of different forms of privileged information using the RetinaNet model on the SODA 1-metre dataset. The comparison includes saliency, depth, fusion, and bounding box mask representations. The bounding box mask yielded the highest improvement in detection accuracy.}
\label{fig:privileged_visual}
\end{figure*}

Selecting effective forms of privileged information for object detection requires cues that meaningfully contribute to both localisation and classification. Cognitive studies suggest that humans rely on \textit{physical reasoning}, such as estimating an object’s center of mass, when recognising objects \cite{object_detection_philospy}. This observation implies that structured spatial signals can enhance detection models.

In deep learning, prior work has investigated auxiliary sources such as saliency \cite{itti, deepgaze, Seychell18} and depth maps \cite{depth_anything, dpt_large}, with saliency shown to correlate more strongly with detection performance \cite{bartolo2024correlationobjectdetectionperformance}. However, directly adding these signals as input channels has yielded limited performance gains. Building on these findings, this study systematically evaluates multiple forms of privileged information within the teacher–student training framework. Both saliency- and depth-based representations were explored for their potential to enhance detection accuracy, yet their contributions remained modest. Among the investigated alternatives, the previously proposed \textit{bounding box mask} \cite{Euvip_Paper} achieved the highest improvement. Figure~\ref{fig:privileged_visual} illustrates the different forms of privileged information alongside corresponding teacher model performance. 

The mask formulation effectively guides the network’s attention toward object regions by embedding both localisation and class cues within a single, structured representation. Each mask image consists of a black background with bounding boxes filled using grayscale values proportional to their class labels. This compact yet informative representation provided the best balance between simplicity, interpretability, and overall detection accuracy.

The mask is generated using ground-truth annotations available only during training, thereby satisfying the LUPI condition. Bounding boxes are drawn in descending size order to minimize occlusion. Although polygonal and segmentation masks \cite{mask_to_annotation, maskrcnn, sam} were also considered, bounding box masks were ultimately preferred for their simplicity, consistency with existing datasets, and stable performance across experiments.

\section{Implementation}
Having outlined the theoretical basis of the proposed LUPI-based detection framework, the next section describes its practical implementation and training setup.

\subsection{Teacher–Student Framework for LUPI}

To leverage the privileged information introduced in the previous section, a teacher–student framework was implemented (see Figure~\ref{fig:lupi_architecture_final}). The teacher network receives RGB images together with the additional privileged input, such as bounding box masks. To accommodate this extra channel, the teacher’s input layer is extended to four channels, with the added weights initialized using Kaiming Normal initialization \cite{kaiming}, while the remaining layers retain pre-trained COCO weights. This adaptation enables the teacher to exploit richer feature representations without modifying the overall architecture.

The student network processes only RGB images but otherwise mirrors the teacher architecture. Its training objective combines detection losses with a knowledge transfer term (see \eqref{eq:lupi-loss}), computed by comparing the student’s and teacher’s feature representations at the final backbone layer using \eqref{eq:cosine-distance}. A weighting parameter \(\alpha\) controls the balance between direct supervision and teacher guidance. During inference, the student operates exclusively on RGB inputs while still benefiting from the knowledge transferred from the teacher.

\subsection{Object Detection Models and Training Protocol}

Building on the teacher–student framework outlined in the previous section, all models were implemented using open-source architectures from the \verb|torchvision|\footnote{\protect\url{https://docs.pytorch.org/vision/main/models.html\#object-detection}} library. The complete training pipeline is publicly available on GitHub\footnote{\protect\url{https://github.com/mbar0075/lupi-for-object-detection}}.

Five object detection models were selected for evaluation: Faster R-CNN \cite{fasterrcnn}, SSD \cite{ssd}, RetinaNet \cite{retinanet}, SSDLite \cite{ssdlite}, and FCOS \cite{fcos}. These architectures cover both one-stage and two-stage detection paradigms and represent a diverse range of computational complexities. The teacher networks were adapted to accept an additional privileged input channel, while the student networks retained standard RGB inputs. Following \eqref{eq:lupi-loss}, knowledge transfer was performed using features from the final backbone layer, which captures semantically rich representations \cite{lab2wild}. Specifically, the last convolutional layer before the FPN was used for Faster R-CNN, FCOS, and RetinaNet, while for SSD and SSDLite, the final convolutional layer before the auxiliary heads was selected (see Figure~\ref{fig:lupi_architecture_final}). This design ensures that performance improvements can be attributed to the integration of privileged information and knowledge transfer, rather than architectural changes.

To maintain consistency across experiments, identical training, preprocessing, and postprocessing procedures were applied to all models. Models were trained for 100 epochs using the Adam optimizer with a fixed learning rate of $1\times10^{-3}$, employing early stopping and checkpointing based on validation loss. Input images, including privileged channels, were normalised using min-max scaling, resized to $800\times800$ pixels, and standardised per channel to zero mean and unit variance. Non-maximum suppression with an IoU threshold of $0.5$ was applied to final predictions to remove redundant detections. 

By standardizing the architectures, training setup, and preprocessing pipelines, this implementation isolates the impact of privileged information and knowledge transfer, ensuring a fair and interpretable evaluation of their contribution to object detection performance.

\section{Evaluation Strategy}
This section presents the experimental evaluation of the proposed LUPI-based object detection framework. It outlines the datasets, metrics, and experimental procedures used to assess the robustness of the approach, followed by a performance analysis across different models and conditions.


\subsection{Datasets and Metrics}
\label{sec:5_evaluation_strategy}
Having defined the LUPI framework and integrated privileged information into the teacher–student setup, the evaluation focused on UAV-based litter detection \cite{soda_dataset, plastopol}---a challenging and practical application due to small object sizes, complex backgrounds, and high scene variability. Publicly available datasets, including SODA \cite{soda_dataset}, BDW \cite{bdwdataset}, and UAVVaste \cite{uavvaste}, were selected for their high-quality annotations and real-world relevance. Subsets of SODA were used for within-dataset experiments to analyze model performance in a controlled setting, while cross-dataset evaluations on BDW and UAVVaste assessed generalization to unseen environments. Additionally, the Pascal VOC 2012 dataset \cite{pascal-voc-2012} was included to evaluate the general applicability of the proposed approach across a broader range of object categories. 

For each model architecture, baseline RGB-only detectors were compared against their LUPI teacher–student counterparts. Teacher networks were also evaluated separately to confirm the contribution of privileged information during training. The study further examined runtime performance, assessing whether student models improved detection accuracy without incurring additional inference cost. Ablation experiments were conducted to investigate the impact of the loss balancing parameter $\alpha$, while qualitative analysis used Grad-CAM visualizations \cite{gradcam} to inspect model attention. Evaluation metrics included standard object detection measures—mAP, precision, recall, F1 score, and mAR along with COCO-style metrics \cite{coco} to assess detection quality across different object scales. This evaluation strategy enabled a consistent and controlled analysis of how privileged information enhances accuracy, generalization, and efficiency across diverse detection models.

\subsection{Within- and Cross-Dataset Experiments}

Within-dataset experiments focused on UAV-based litter detection using subsets of the SODA dataset to evaluate model performance under controlled conditions. Three scenarios were explored: (i) binary litter detection at 1-metre altitude without tiling, (ii) binary detection across multiple altitudes with 3$\times$3 tiling, and (iii) multi-label detection across altitudes also with 3$\times$3 tiling. For each scenario, the five selected object detection architectures were trained as both teacher and student models.  The parameter \(\alpha\) was varied from 0 to 1 in steps of 0.25, following the methodology of \cite{lab2wild}, to analyze the effect of teacher supervision strength. This experimental design enabled a systematic evaluation of how the LUPI framework enhances student performance.


Cross-dataset experiments evaluated the generalization capacity of models trained on SODA when applied to other litter datasets. For BDW, models trained on SODA at an altitude of 1-metre were directly tested without retraining, while UAVVaste evaluations used models trained on 3$\times$3 tiled SODA images across multiple altitudes. All experiments focused on binary detection, enabling analysis of how effectively LUPI-trained students adapt to unseen environments and varying object distributions. These evaluations also examined runtime considerations, emphasizing performance gains achieved without increasing model size or inference time.

\definecolor{colorAP}{HTML}{FFFFC7} 
\definecolor{colorAR}{HTML}{E5B0FC} 


\definecolor{soda1mText}{HTML}{000000}       
\definecolor{sodaTiledBinaryText}{HTML}{000000} 
\definecolor{sodaTiledMultiText}{HTML}{000000}  
\definecolor{bdwText}{HTML}{000000}            
\definecolor{uavText}{HTML}{000000}            
\definecolor{vocText}{HTML}{000000}                

\definecolor{retinaText}{HTML}{000080} 
\definecolor{fcosText}{HTML}{006400}   
\definecolor{fasterText}{HTML}{0000CD} 
\definecolor{ssdText}{HTML}{8B0000}    
\definecolor{ssdliteText}{HTML}{4B0082} 

\begin{table*}[!t]
\centering
\caption{Comparison of teacher model performance across all experiments using COCO metrics (2 decimal places). Includes within-dataset, cross-dataset, and Pascal VOC 2012 evaluations. Faster R-CNN shows the highest average performance, with RetinaNet and FCOS performing similarly, while SSD and SSDLite exhibit lower results. Note that the SODA 1-metre subset contains no small objects. For cross-dataset evaluations, privileged information was also generated, and the teacher models were evaluated accordingly.}
\label{tab:performance_2dp}
\begin{adjustbox}{max width=\textwidth, center}
\renewcommand{\arraystretch}{1.2}
\begin{tabular}{l c ccc ccc ccc ccc}
\toprule
\multicolumn{1}{c}{\multirow{2}{*}{Model}}  &
\multirow{2}{*}{Dataset} &
\multicolumn{3}{c}{mAP} &
\multicolumn{3}{c}{mAP} &
\multicolumn{3}{c}{mAR} &
\multicolumn{3}{c}{mAR} \\
\cmidrule(lr){3-5} \cmidrule(lr){6-8} \cmidrule(lr){9-11} \cmidrule(lr){12-14}
& & mAP & @50 & @75 &
@Small & @Medium & @Large &
@1 & @10 & @100 &
@Small & @Medium & @Large \\
\midrule
\textcolor{retinaText}{RetinaNet} & \multirow{5}{*}{\textcolor{soda1mText}{SODA at 1-metre}} & \cellcolor{colorAP}0.94 & \cellcolor{colorAP}0.98 & \cellcolor{colorAP}0.96 & – & 0.93 & 0.95 & \cellcolor{colorAR}0.63 & \cellcolor{colorAR}0.96 & \cellcolor{colorAR}0.96 & – & 0.94 & 0.97 \\
\textcolor{fcosText}{FCOS}  &  & \cellcolor{colorAP}0.96 & \cellcolor{colorAP}0.98 & \cellcolor{colorAP}0.97 & – & 0.93 & \textbf{0.97} & \cellcolor{colorAR}\textbf{0.63} & \cellcolor{colorAR}0.97 & \cellcolor{colorAR}0.97 & – & 0.94 & \textbf{0.98} \\
\textcolor{fasterText}{Faster R-CNN} & & \cellcolor{colorAP}\textbf{0.96} & \cellcolor{colorAP}\textbf{0.99} & \cellcolor{colorAP}\textbf{0.98} & – & \textbf{0.98} & 0.96 & \cellcolor{colorAR}0.63 & \cellcolor{colorAR}\textbf{0.98} & \cellcolor{colorAR}\textbf{0.98} & – & \textbf{0.99} & 0.97 \\
\textcolor{ssdText}{SSD} & & \cellcolor{colorAP}0.78 & \cellcolor{colorAP}0.96 & \cellcolor{colorAP}0.94 & – & 0.78 & 0.78 & \cellcolor{colorAR}0.54 & \cellcolor{colorAR}0.81 & \cellcolor{colorAR}0.81 & – & 0.82 & 0.81 \\
\textcolor{ssdliteText}{SSDLite} & & \cellcolor{colorAP}0.61 & \cellcolor{colorAP}0.73 & \cellcolor{colorAP}0.72 & – & 0.00 & 0.73 & \cellcolor{colorAR}0.48 & \cellcolor{colorAR}0.63 & \cellcolor{colorAR}0.63 & – & 0.00 & 0.77 \\
\hline
\textcolor{retinaText}{RetinaNet} & \multirow{5}{*}{\textcolor{sodaTiledBinaryText}{SODA Tiled Binary}} & \cellcolor{colorAP}0.90 & \cellcolor{colorAP}0.95 & \cellcolor{colorAP}0.94 & 0.78 & 0.98 & 0.97 & \cellcolor{colorAR}0.34 & \cellcolor{colorAR}0.83 & \cellcolor{colorAR}0.91 & 0.84 & 0.98 & 0.98 \\
\textcolor{fcosText}{FCOS}  & & \cellcolor{colorAP}0.89 & \cellcolor{colorAP}0.94 & \cellcolor{colorAP}0.93 & 0.80 & 0.95 & 0.97 & \cellcolor{colorAR}0.34 & \cellcolor{colorAR}0.82 & \cellcolor{colorAR}0.90 & 0.83 & 0.97 & 0.98 \\
\textcolor{fasterText}{Faster R-CNN} & & \cellcolor{colorAP}\textbf{0.96} & \cellcolor{colorAP}\textbf{0.99} & \cellcolor{colorAP}\textbf{0.98} & \textbf{0.92} & \textbf{0.99} & \textbf{0.99} & \cellcolor{colorAR}\textbf{0.35} & \cellcolor{colorAR}\textbf{0.87} & \cellcolor{colorAR}\textbf{0.97} & \textbf{0.94} & \textbf{0.99} & \textbf{0.99} \\
\textcolor{ssdText}{SSD} & & \cellcolor{colorAP}0.49 & \cellcolor{colorAP}0.62 & \cellcolor{colorAP}0.59 & 0.19 & 0.77 & 0.75 & \cellcolor{colorAR}0.27 & \cellcolor{colorAR}0.51 & \cellcolor{colorAR}0.51 & 0.21 & 0.80 & 0.80 \\
\textcolor{ssdliteText}{SSDLite} & & \cellcolor{colorAP}0.18 & \cellcolor{colorAP}0.23 & \cellcolor{colorAP}0.19 & 0.00 & 0.05 & 0.80 & \cellcolor{colorAR}0.17 & \cellcolor{colorAR}0.19 & \cellcolor{colorAR}0.19 & 0.01 & 0.07 & 0.83 \\
\hline
\textcolor{retinaText}{RetinaNet} & \multirow{5}{*}{\textcolor{sodaTiledMultiText}{SODA Tiled Multi-label}} & \cellcolor{colorAP}0.88 & \cellcolor{colorAP}0.92 & \cellcolor{colorAP}0.91 & 0.75 & 0.98 & 0.98 & \cellcolor{colorAR}0.66 & \cellcolor{colorAR}0.89 & \cellcolor{colorAR}0.89 & 0.77 & 0.98 & 0.99 \\
\textcolor{fcosText}{FCOS}  & & \cellcolor{colorAP}0.91 & \cellcolor{colorAP}0.95 & \cellcolor{colorAP}0.94 & 0.83 & 0.97 & 0.97 & \cellcolor{colorAR}0.68 & \cellcolor{colorAR}0.92 & \cellcolor{colorAR}0.92 & 0.85 & 0.98 & 0.98 \\
\textcolor{fasterText}{Faster R-CNN} & & \cellcolor{colorAP}\textbf{0.95} & \cellcolor{colorAP}\textbf{0.99} & \cellcolor{colorAP}\textbf{0.98} & \textbf{0.91} & \textbf{0.98} & \textbf{0.98} & \cellcolor{colorAR}\textbf{0.70} & \cellcolor{colorAR}\textbf{0.96} & \cellcolor{colorAR}\textbf{0.96} & \textbf{0.93} & \textbf{0.99} & \textbf{0.99} \\
\textcolor{ssdText}{SSD} & & \cellcolor{colorAP}0.36 & \cellcolor{colorAP}0.49 & \cellcolor{colorAP}0.45 & 0.15 & 0.55 & 0.55 & \cellcolor{colorAR}0.33 & \cellcolor{colorAR}0.41 & \cellcolor{colorAR}0.41 & 0.16 & 0.59 & 0.62 \\
\textcolor{ssdliteText}{SSDLite} & & \cellcolor{colorAP}0.11 & \cellcolor{colorAP}0.13 & \cellcolor{colorAP}0.13 & 0.00 & 0.00 & 0.46 & \cellcolor{colorAR}0.13 & \cellcolor{colorAR}0.13 & \cellcolor{colorAR}0.13 & 0.00 & 0.00 & 0.54 \\
\hline
\textcolor{retinaText}{RetinaNet} & \multirow{5}{*}{\textcolor{bdwText}{BDW}} & \cellcolor{colorAP}0.46 & \cellcolor{colorAP}0.96 & \cellcolor{colorAP}0.32 & 0.00 & 0.35 & 0.52 & \cellcolor{colorAR}0.38 & \cellcolor{colorAR}0.54 & \cellcolor{colorAR}0.54 & 0.00 & 0.40 & \textbf{0.59} \\
\textcolor{fcosText}{FCOS}  & & \cellcolor{colorAP}0.49 & \cellcolor{colorAP}0.96 & \cellcolor{colorAP}0.43 & 0.10 & 0.37 & 0.55 & \cellcolor{colorAR}0.39 & \cellcolor{colorAR}0.56 & \cellcolor{colorAR}0.56 & 0.10 & 0.43 & 0.61 \\
\textcolor{fasterText}{Faster R-CNN} & & \cellcolor{colorAP}0.48 & \cellcolor{colorAP}\textbf{0.97} & \cellcolor{colorAP}0.34 & \textbf{0.20} & 0.40 & 0.52 & \cellcolor{colorAR}0.38 & \cellcolor{colorAR}0.54 & \cellcolor{colorAR}0.54 & \textbf{0.20} & 0.41 & 0.59 \\
\textcolor{ssdText}{SSD} & & \cellcolor{colorAP}\textbf{0.55} & \cellcolor{colorAP}0.95 & \cellcolor{colorAP}\textbf{0.62} & 0.00 & \textbf{0.45} & \textbf{0.58} & \cellcolor{colorAR}\textbf{0.43} & \cellcolor{colorAR}\textbf{0.59} & \cellcolor{colorAR}\textbf{0.59} & 0.00 & \textbf{0.48} & 0.64 \\
\textcolor{ssdliteText}{SSDLite} & & \cellcolor{colorAP}0.23 & \cellcolor{colorAP}0.37 & \cellcolor{colorAP}0.27 & 0.00 & 0.01 & 0.31 & \cellcolor{colorAR}0.21 & \cellcolor{colorAR}0.24 & \cellcolor{colorAR}0.24 & 0.00 & 0.01 & 0.34 \\
\hline
\textcolor{retinaText}{RetinaNet} & \multirow{5}{*}{\textcolor{uavText}{UAVVaste}} & \cellcolor{colorAP}0.40 & \cellcolor{colorAP}0.78 & \cellcolor{colorAP}0.37 & 0.31 & 0.72 & \textbf{0.93} & \cellcolor{colorAR}0.13 & \cellcolor{colorAR}0.44 & \cellcolor{colorAR}\textbf{0.47} & \textbf{0.41} & 0.74 & \textbf{0.95} \\
\textcolor{fcosText}{FCOS}  & & \cellcolor{colorAP}0.42 & \cellcolor{colorAP}0.71 & \cellcolor{colorAP}\textbf{0.47} & 0.36 & 0.74 & 0.90 & \cellcolor{colorAR}0.14 & \cellcolor{colorAR}\textbf{0.46} & \cellcolor{colorAR}0.46 & 0.39 & 0.76 & 0.90 \\
\textcolor{fasterText}{Faster R-CNN} & & \cellcolor{colorAP}\textbf{0.44} & \cellcolor{colorAP}\textbf{0.84} & \cellcolor{colorAP}0.44 & \textbf{0.37} & \textbf{0.75} & 0.85 & \cellcolor{colorAR}\textbf{0.15} & \cellcolor{colorAR}0.46 & \cellcolor{colorAR}0.46 & 0.39 & \textbf{0.77} & 0.85 \\
\textcolor{ssdText}{SSD} & & \cellcolor{colorAP}0.13 & \cellcolor{colorAP}0.24 & \cellcolor{colorAP}0.12 & 0.06 & 0.45 & 0.80 & \cellcolor{colorAR}0.08 & \cellcolor{colorAR}0.15 & \cellcolor{colorAR}0.15 & 0.08 & 0.50 & 0.80 \\
\textcolor{ssdliteText}{SSDLite} & & \cellcolor{colorAP}0.01 & \cellcolor{colorAP}0.03 & \cellcolor{colorAP}0.01 & 0.00 & 0.06 & 0.20 & \cellcolor{colorAR}0.01 & \cellcolor{colorAR}0.01 & \cellcolor{colorAR}0.01 & 0.00 & 0.07 & 0.20 \\
\hline
\textcolor{retinaText}{RetinaNet} & \multirow{5}{*}{\textcolor{vocText}{Pascal VOC 2012}} & \cellcolor{colorAP}0.77 & \cellcolor{colorAP}0.86 & \cellcolor{colorAP}0.79 & 0.28 & 0.63 & 0.80 & \cellcolor{colorAR}0.60 & \cellcolor{colorAR}0.81 & \cellcolor{colorAR}0.81 & 0.30 & 0.67 & 0.84 \\
\textcolor{fcosText}{FCOS}  & & \cellcolor{colorAP}\textbf{0.80} & \cellcolor{colorAP}\textbf{0.88} & \cellcolor{colorAP}\textbf{0.82} & \textbf{0.56} & \textbf{0.67} & \textbf{0.83} & \cellcolor{colorAR}\textbf{0.61} & \cellcolor{colorAR}\textbf{0.84} & \cellcolor{colorAR}\textbf{0.84} & \textbf{0.57} & 0.72 & \textbf{0.86} \\
\textcolor{fasterText}{Faster R-CNN} & & \cellcolor{colorAP}0.77 & \cellcolor{colorAP}0.91 & \cellcolor{colorAP}0.82 & 0.51 & 0.66 & 0.79 & \cellcolor{colorAR}0.59 & \cellcolor{colorAR}0.82 & \cellcolor{colorAR}0.82 & 0.56 & \textbf{0.72} & 0.85 \\
\textcolor{ssdText}{SSD} & & \cellcolor{colorAP}0.42 & \cellcolor{colorAP}0.56 & \cellcolor{colorAP}0.49 & 0.00 & 0.06 & 0.48 & \cellcolor{colorAR}0.41 & \cellcolor{colorAR}0.48 & \cellcolor{colorAR}0.48 & 0.00 & 0.07 & 0.56 \\
\textcolor{ssdliteText}{SSDLite} & & \cellcolor{colorAP}0.49 & \cellcolor{colorAP}0.61 & \cellcolor{colorAP}0.54 & 0.00 & 0.00 & 0.58 & \cellcolor{colorAR}0.46 & \cellcolor{colorAR}0.55 & \cellcolor{colorAR}0.55 & 0.00 & 0.00 & 0.65 \\
\bottomrule
\end{tabular}
\renewcommand{\arraystretch}{1}
\end{adjustbox}
\end{table*}

\subsection{Pascal VOC 2012 Experiment}

To assess the broader applicability of the proposed LUPI framework, experiments were conducted on the Pascal VOC 2012 dataset, which includes multi-label detection across 20 diverse object categories. This evaluation examined whether student models could effectively leverage teacher guidance in complex scenes containing multiple objects and varying scales, extending the analysis beyond UAV-specific litter detection. Baseline RGB-only models, LUPI student models, and teacher networks were compared using identical architectures and training protocols, ensuring a controlled assessment of generalization across a more heterogeneous set of classes.

\subsection{Ablation Study on Teacher–Student Balance}

Ablation studies investigated the effect of the balancing parameter $\alpha$ on student performance, which regulates the contribution of teacher supervision relative to ground-truth labels. Values of  ranging from 0 to 1, in increments of 0.25 as in \cite{lab2wild}, were tested across the SODA dataset (binary and multi-label scenarios) and Pascal VOC 2012 to determine optimal weighting for different tasks. These experiments provided insight into how the degree of teacher reliance influences learning dynamics, guiding the selection of $\alpha$ values that maximize performance while avoiding excessive dependence on privileged information.

\section{Results and Discussion}
Based on the evaluation setups described above, results are presented collectively to enable direct comparison across experiments, covering performance, ablation, interpretability, and efficiency.

\subsection{Teacher Model Performance}
Having established the optimal form of privileged information (Figure~\ref{fig:privileged_visual}), it is essential to assess its impact across different architectures and datasets. Table~\ref{tab:performance_2dp} summarizes teacher model performance for all experiments---within-dataset, cross-dataset, and Pascal VOC 2012---covering the five selected detection architectures. The results indicate that incorporating informative privileged input substantially improves teacher accuracy, with strict mAP and mAR values approaching 1, demonstrating high reliability. Although the improvement is less pronounced for small objects, reflecting the difficulty of this category, performance for medium and large objects remains consistently strong, suggesting that privileged signals help the model focus on more easily detectable targets. While performance decreases slightly under more challenging conditions, such as cross-dataset generalization and multi-label detection, teacher models still maintain robust accuracy. Overall, Faster R-CNN achieves the highest average mAP, followed closely by RetinaNet and FCOS, whereas SSD and SSDLite exhibit comparatively lower performance.

\subsection{Baseline vs. Student Model Comparison}

\begin{figure*}[ht]
\centering
\includegraphics[width=1\textwidth]{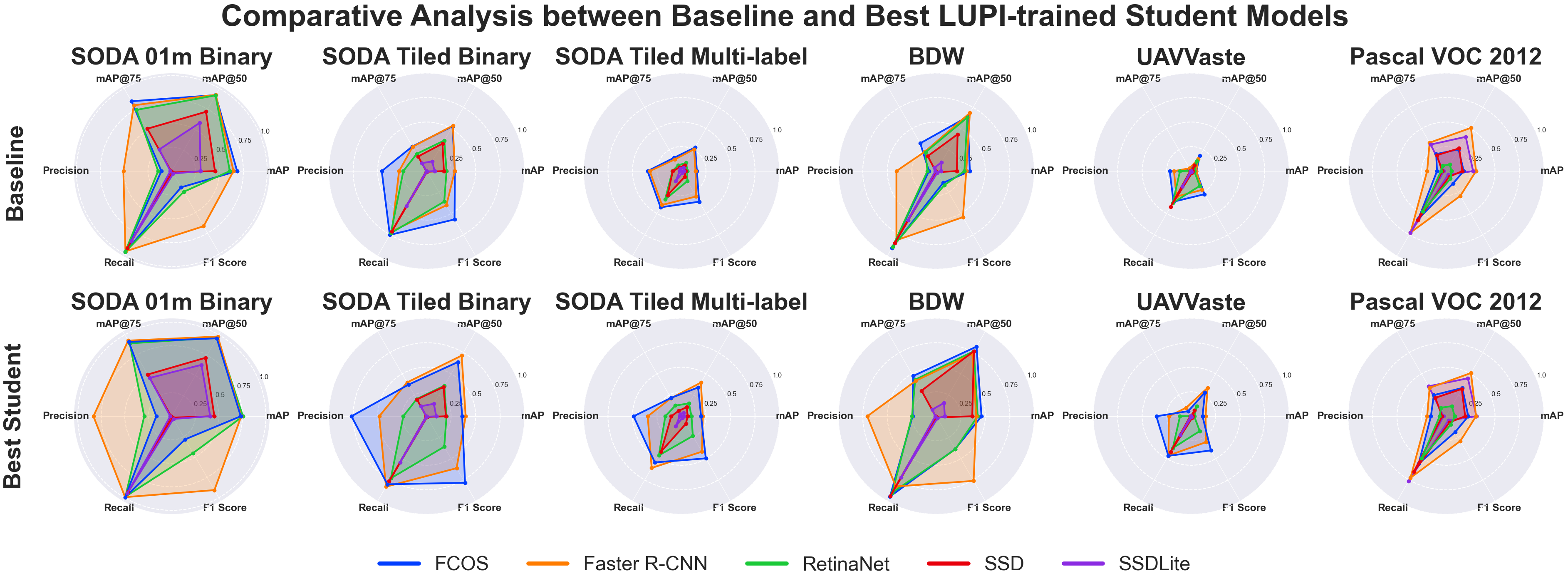}
\caption{Comparative analysis of baseline and best LUPI-trained student models across all datasets for the five architectures, shown as a multi-radar graph. The figure highlights notable improvements in strict mAP and F1 score, with the largest boosts observed in within-dataset evaluations, while other datasets show smaller yet meaningful improvements using identical architectures.}
\label{fig:evaluation_radar}
\end{figure*}

Following the teacher model evaluation, we assess the performance of LUPI-trained student models relative to their baseline RGB-only counterparts. Figure~\ref{fig:evaluation_radar} presents these comparisons across all datasets and detection architectures. Overall, student models show consistent gains over the baselines, particularly in strict mAP and F1 score metrics. The most substantial improvements are observed in within-dataset UAV litter detection, with smaller yet meaningful gains in cross-dataset evaluations. Faster R-CNN, FCOS, and RetinaNet benefit most from teacher guidance in UAV-based scenarios, whereas SSD and SSDLite exhibit clearer improvements on Pascal VOC. Although relative gains diminish in more demanding tasks---such as multi-label detection and cross-dataset generalization---the results confirm that LUPI effectively enhances student performance across architectures and domains. 

\subsection{The Effect of Balancing the $\alpha$ Parameter}

\begin{figure*}[ht]
\centering
\includegraphics[width=1\textwidth]{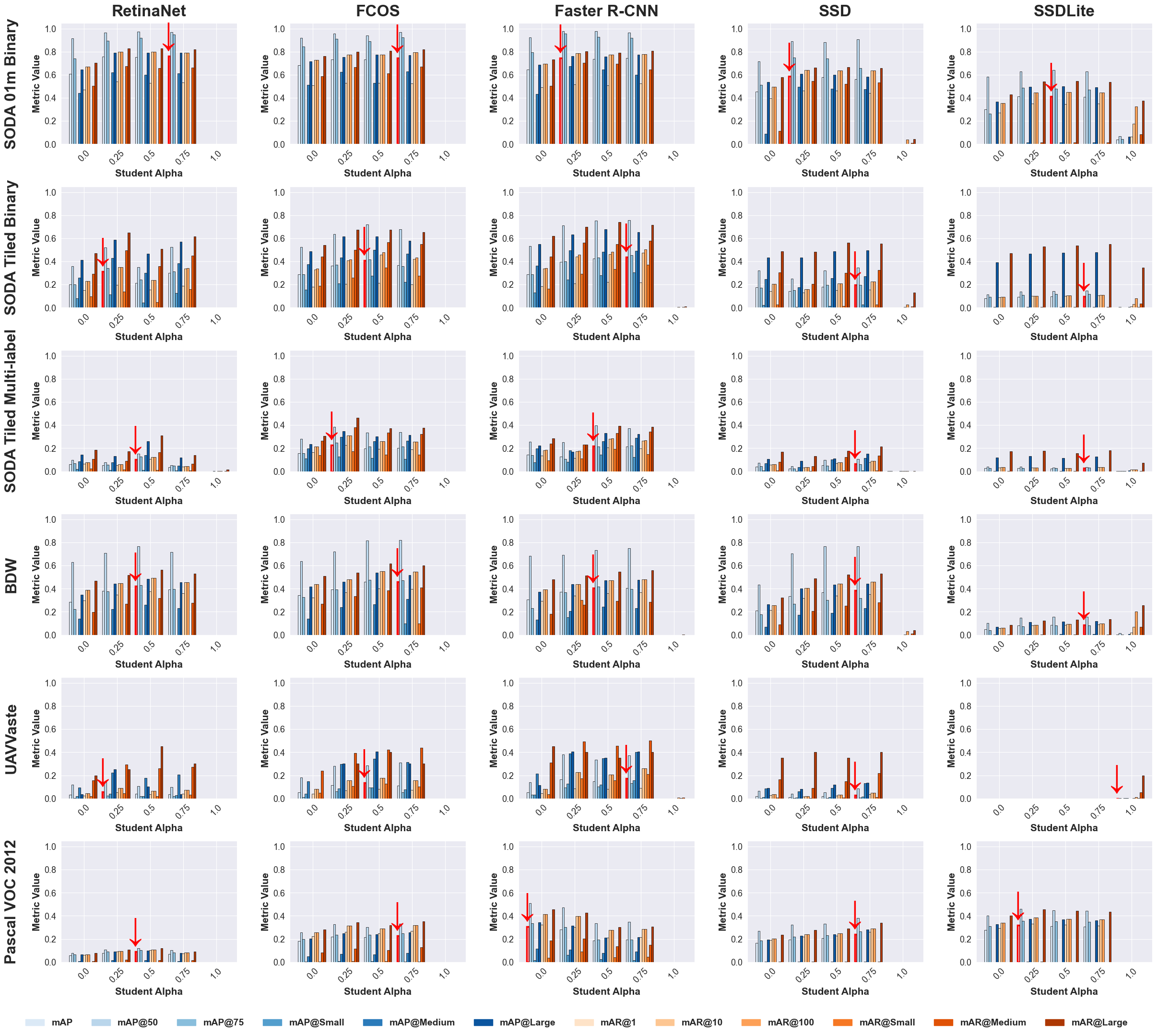}
\caption{Ablation study results across all datasets and experiments using COCO metrics. Baseline model corresponds to $\alpha = 0$; other lines represent student models. Red downward arrows indicate top performance in the strict map@50--95 metric. Best results are generally observed for $\alpha = 0.25$ and 0.5, with $\alpha = 0.75$ also performing well, while $\alpha = 1$ shows lower average performance.}
\label{fig:all results}
\end{figure*}

While the teacher models exhibited high accuracy and the student models consistently outperformed their baselines, it remains important to analyze how varying reliance on teacher guidance affects student learning. This dependency is governed by the parameter $\alpha$, which controls the weight of the teacher’s contribution during knowledge transfer. Experiments were conducted with $\alpha \in \{0, 0.25, 0.5, 0.75, 1\}$, where $\alpha = 0$ corresponds to the baseline and $\alpha = 1$ to full teacher supervision. Figure~\ref{fig:all results} summarizes the results across all datasets and architectures using COCO metrics, with red downward arrows indicating top performance in strict mAP@50--95. Intermediate values of $\alpha$ (0.25 and 0.5) generally yield the best performance, balancing learning from ground-truth labels and teacher knowledge, while $\alpha = 0.75$ occasionally performs well and $\alpha = 1$ tends to underperform. These trends are consistent across datasets and models.

Smaller objects continue to present challenges, showing limited improvement, whereas medium and large objects benefit more significantly, reflecting the richer semantic features transferred from the teacher. It is also observed that SSD and SSDLite perform comparatively worse on certain datasets, consistent with their architectural limitations. In the case of Faster R-CNN, only the Pascal VOC baseline outperformed its student counterpart, likely because the teacher’s additional region proposals introduced ambiguity in supervision limiting the student’s gains, though this difference remains marginal.

\subsection{Interpretability Analysis}

\begin{figure*}[ht]
\centering
\includegraphics[width=0.9\textwidth]{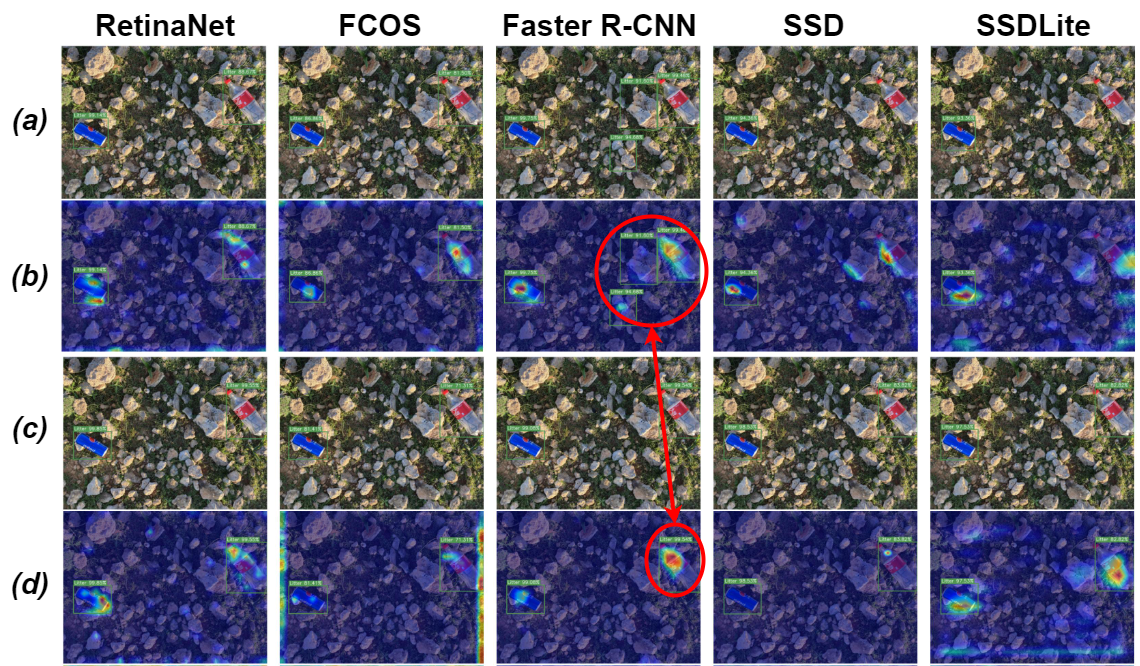}
\caption{Visual comparison of model predictions and interpretability results on the SODA 1-metre dataset experiment. (a) Baseline detection results. (b) Baseline Grad-CAM visualisation. (c) Best LUPI-trained student detection results. (d) Best Student Grad-CAM visualisation. The LUPI-trained student produces more accurate litter predictions than the baseline. For the Grad-CAM visualisations applied to the respective distillation layers, the student’s attention is more concentrated on litter objects, whereas the baseline exhibits more diffuse activation across the background.}
\label{fig:gradcam_journal}
\end{figure*}

To further understand the performance improvements observed in LUPI-trained student models, we performed interpretability analysis using Grad-CAM visualisations \cite{gradcam} on the final backbone layer. Figure~\ref{fig:gradcam_journal} shows results for the SODA 1-metre dataset, comparing baseline and student models. The visualisations show that LUPI-trained student models focus sharply on litter objects, producing higher-confidence detections with fewer misclassifications, whereas the baseline model's attention is less concentrated and often highlights irrelevant areas in the background. This targeted attention aligns with the improvements observed in strict mAP and F1 score, indicating that the student models are not only performing better quantitatively but also learning more meaningful and task-relevant feature representations. Similar patterns were observed across the other datasets, though these results are omitted for brevity, indicating that the effect is consistent.

\subsection{Performance and Efficiency Analysis}

\begin{figure}[ht]
    \centering
    \includegraphics[width=1\columnwidth]{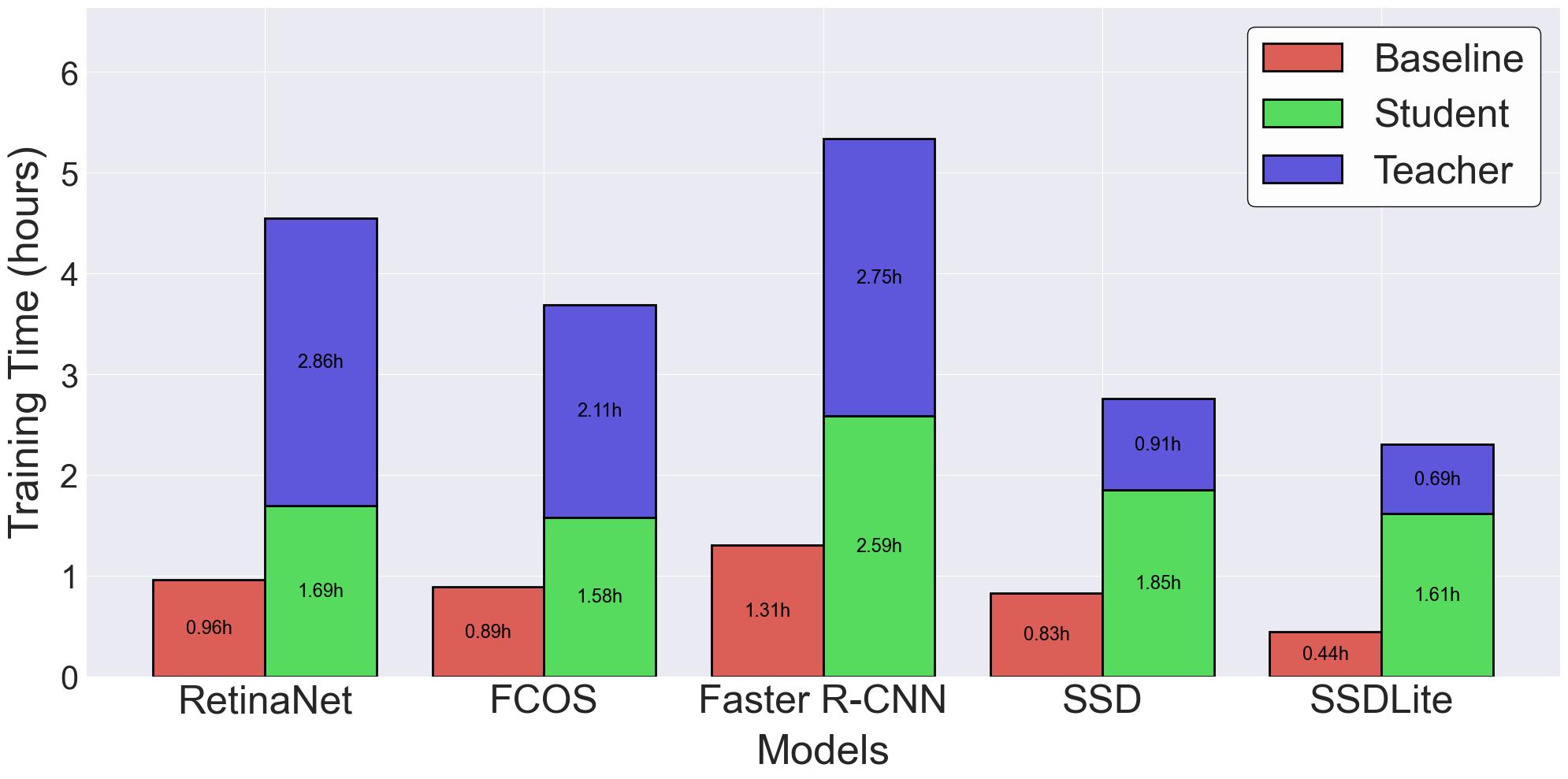}
    \caption{Comparison of training times on the Pascal VOC 2012 dataset, highlighting the increased duration for LUPI teacher–student training.}
    \label{fig:pascal_voc_training_time}
\end{figure}

While we have consistently shown that the proposed LUPI-based object detection approach improves accuracy, one might question whether these results come at a substantial computational cost. A primary limitation is the increased training time, as both a teacher and a student model must be trained, effectively doubling the workload, as shown in Figure~\ref{fig:pascal_voc_training_time}. However, inference is typically performed far more frequently than training in practical deployments. Table~\ref{tab:model_configs_pascal_voc} shows that the baseline and LUPI-trained student models are nearly identical in size, number of parameters, GFLOPS, and FPS, with only minor variations in speed that were consistent across multiple runs. This demonstrates that while training requires more time, inference speed, model size, and efficiency remain unaffected, ensuring that the improved accuracy of LUPI-trained students can be fully utilised in deployment. For brevity, only Pascal VOC 2012 results are shown, though similar trends were observed across other datasets.

\begin{table}[ht]
    \centering
    \caption{Runtime comparison of baseline and student models on Pascal VOC 2012, showing model type, size, parameters, GFLOPS, and FPS. Performance improvements for students come with no additional inference cost.}
     \begin{adjustbox}{width=\columnwidth}
    \begin{tabular}{llcccc}
        \toprule
        \textbf{Model} & \textbf{Type} & \textbf{Size (MB)} & \textbf{Parameters (M)} & \textbf{GFLOPS} & \textbf{FPS} \\
        \midrule
        \multirow{5}{*}{\textbf{Baseline}} 
            & RetinaNet     & 124.22 & 32.56 & 265.10 & 39.74 \\
            & FCOS          & 122.48 & 32.11 & 252.94 & 39.55 \\
            & Faster R-CNN  & 157.92 & 41.40 & 268.75 & 30.66 \\
            & SSD           & 100.27 & 26.29 & 62.94 & 67.44 \\
            & SSDLite       & 9.42   & 2.47 & 0.95 & 36.74 \\
        \midrule
        \multirow{5}{*}{\textbf{Student}} 
            & RetinaNet     & 124.22 & 32.56 & 265.10 & 38.00 \\
            & FCOS          & 122.48 & 32.11 & 252.94 & 34.65 \\
            & Faster R-CNN  & 157.92 & 41.40 & 268.75 & 30.39 \\
            & SSD           & 100.27 & 26.29 & 62.94 & 67.67 \\
            & SSDLite       & 9.42   & 2.47 & 0.95 & 36.75 \\
        \bottomrule
    \end{tabular}
    \end{adjustbox}
    \label{tab:model_configs_pascal_voc}
\end{table}

\subsection{Discussion}
Object detection is a multifaceted problem, as different architectures handle localisation and classification in distinct ways. Two-stage detectors with region proposal networks and feature pyramids leverage spatial and semantic cues, while single-stage models with auxiliary layers or lightweight designs depend more on end-to-end feature extraction, affecting how they respond to additional guidance. Across our experiments, integrating privileged information through the LUPI framework consistently improved student learning, although the magnitude and nature of these improvements varied across models and datasets. Faster R-CNN, FCOS, and RetinaNet were more effective in UAV litter detection, reflecting their ability to utilise spatial context, while SSD and SSDLite performed comparatively better on Pascal VOC tasks, highlighting differences in feature aggregation and receptive fields. Ablation studies show that moderate teacher weighting supports student learning by balancing reliance on ground-truth supervision with teacher guidance, whereas excessive dependence can occasionally confuse the student in complex multi-label scenarios. Grad-CAM visualisations indicate that LUPI-trained students focus more sharply on relevant objects, producing more discriminative and semantically coherent representations, while baseline models show more diffuse attention. Importantly, these improvements are achieved with minimal architectural changes, demonstrating that the framework complements the intrinsic characteristics of each model, and inference speed and efficiency remain consistent. Overall, the results illustrate a nuanced interaction between architecture, dataset characteristics, and teacher guidance, emphasising that the benefits of privileged information depend on both model design and task complexity, with LUPI serving as an augmenter of the capabilities already present in the underlying model.

\section{Practical Applications}
The LUPI paradigm within object detection, as presented in this study, can be applied to a wide range of object detection and geolocation systems and is especially well-suited for lightweight deployment. By leveraging compact models, it reduces inference costs while maintaining high accuracy, enabling efficient processing even on resource-constrained platforms. This makes the approach suitable for real-world applications that demand fast, reliable, and consistent detection, such as UAV monitoring, traffic analysis, and surveillance systems, where both speed and precision are critical.


\section{Conclusion}

This study investigated the LUPI paradigm within object detection through an extensive series of experiments across multiple architectures and datasets, evaluated using strict COCO metrics. The results consistently demonstrate that incorporating privileged information during training enhances detection accuracy, improving accuracy without increasing model depth, parameter count, or inference time. The student models trained under this paradigm remained identical to their baseline counterparts in architecture and efficiency, yet achieved higher accuracy through the use of additional teacher guidance during training.

Some limitations remain. The generation of privileged information can be affected by overlapping objects of the same category, occlusions from larger bounding boxes, and limited color differentiation within mask representations. Moreover, the need to train both a teacher and a student model introduces a longer training phase compared to conventional single-model setups.

Future avenues extending this work include the integration of the approach with more recent detection architectures such as YOLOv12 \cite{yolov12} and RF-DETR \cite{rf-detr}, the exploration of richer and more diverse forms of privileged information such as semantic maps or attention-based cues, and the adaptation of the framework to related tasks like object segmentation. Overall, the findings of this study reaffirm that LUPI provides a practical and effective strategy for enhancing object detection performance in computationally constrained environments, maintaining identical inference efficiency while achieving higher accuracy.

\bibliographystyle{IEEEtran}
\bibliography{references.bib}

\begin{IEEEbiography}[{\includegraphics[width=1in,height=1.25in,clip,keepaspectratio]{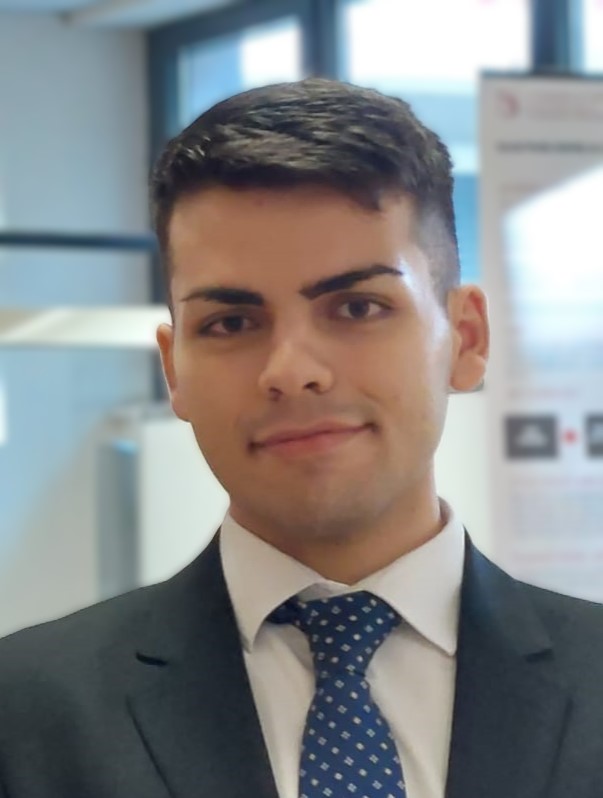}}]{Matthias Bartolo}
received the B.Sc. IT degree (Hons.) in Artificial Intelligence from the University of Malta with First Class Honours, ranking first in his cohort, and was recognised on the Dean’s List for outstanding academic achievement. He subsequently obtained the M.Sc. degree from the University of Malta. His research interests include computer vision, applied machine learning, and the development of AI solutions that assist and support people. He has contributed to several research projects and peer-reviewed publications. In addition, Matthias is actively involved in voluntary initiatives, reflecting a strong commitment to applying his technical expertise in the service of others.
\end{IEEEbiography}

\begin{IEEEbiography}[{\includegraphics[width=1in,height=1.25in,clip,keepaspectratio]{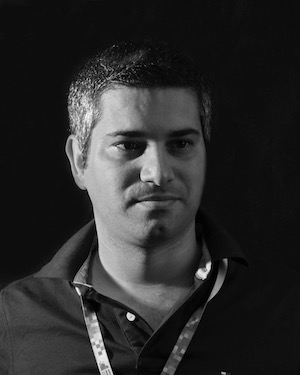}}]{Dylan Seychell} (Senior Member, IEEE)
received his BSc IT (Hons), M.SC. and Ph.D. degrees in computer vision from the University of Malta, Msida, Malta. He is currently a Resident Academic and Senior Lecturer in the Department of Artificial Intelligence at the University of Malta. His research interests include computer vision, applied machine learning, and remote sensing, with a focus on deploying AI systems in operational environments. Dr Seychell serves as the Principal Investigator for several nationally funded research projects focusing on environmental sustainability and media integrity. He is also a Technical Expert with the Malta Digital Innovation Authority (MDIA).
\end{IEEEbiography}

\begin{IEEEbiography}[{\includegraphics[width=1in,height=1.25in,clip,keepaspectratio]{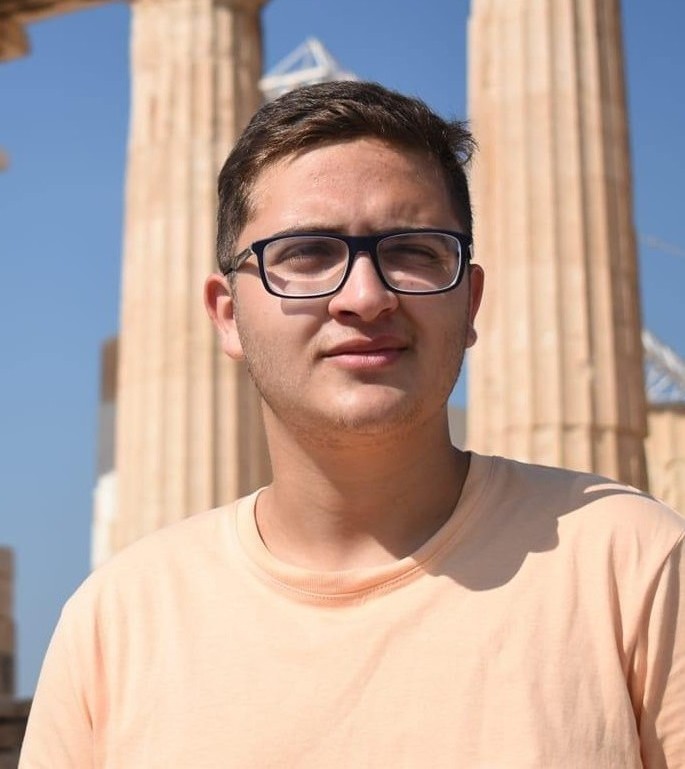}}]{Gabriel Hili} received his M.Sc. degree in applied computer vision in the context of news media from the University of Malta. He is currently a researcher at the University of Malta, where he works on a collaborative research project with the Government of Malta’s Cleansing and Maintenance Division. His research interests include computer vision and language technologies, with a focus on the practical application of AI in public infrastructure and media. Mr Hili also contributes to the academic curriculum by assisting senior faculty in the delivery of lectures and the development of teaching materials.
\end{IEEEbiography}

\begin{IEEEbiography}[{\includegraphics[width=1in,height=1.25in,clip,keepaspectratio]{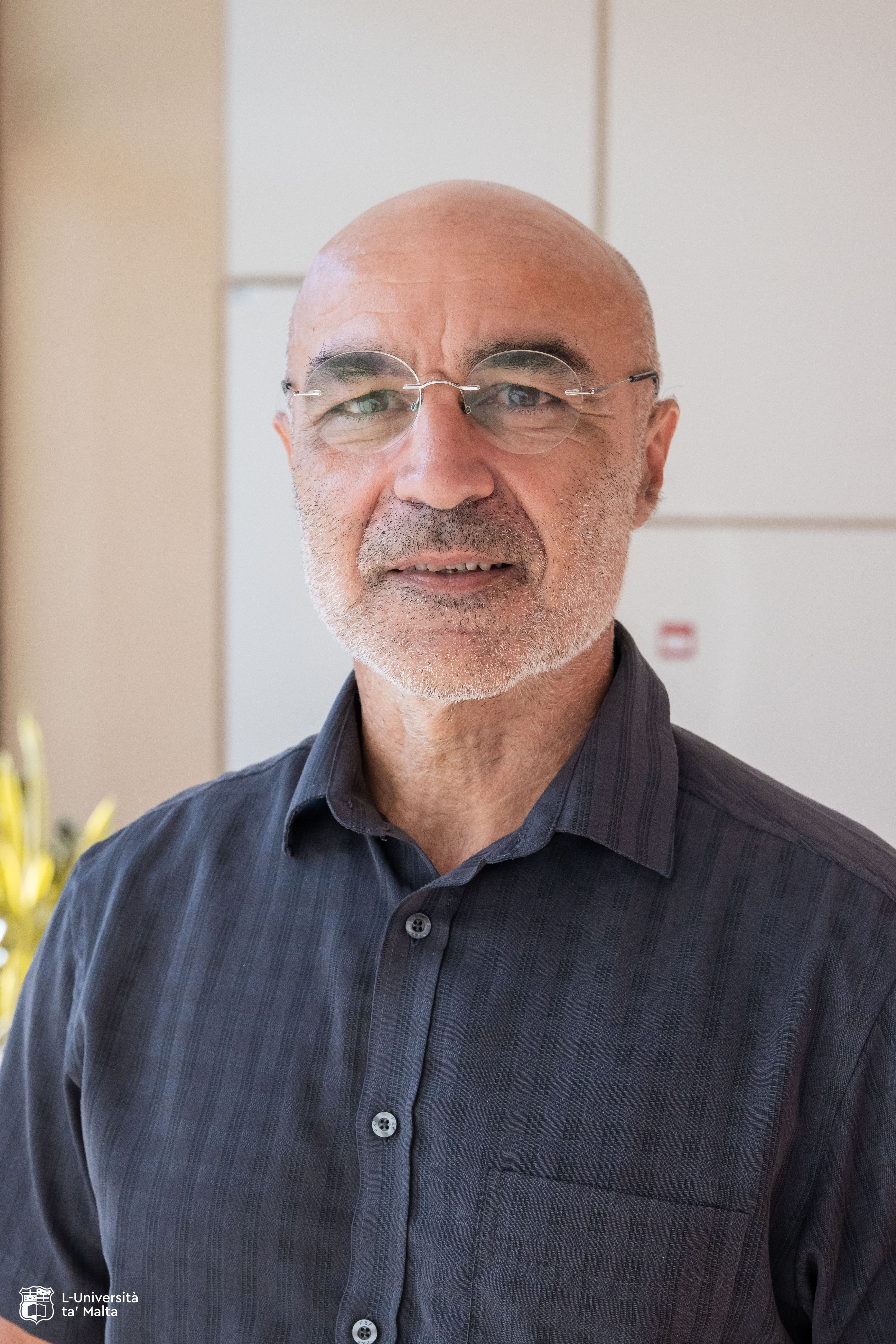}}]{Matthew Montebello} (Senior Member, IEEE)
is an academic and researcher at the University of Malta with over 35 years of experience in higher education. He holds doctorates in Computer Science and Education, with research expertise in personalisation, artificial intelligence, and digital education. His work focuses on AI in higher education, generative AI, ePortfolios, and ethical, human-centred educational technologies. He has led and coordinated multiple nationally and internationally funded projects, published with leading academic publishers, and received awards for innovation and research. He is actively involved in curriculum design, doctoral supervision, and institutional AI strategy.
\end{IEEEbiography}

\begin{IEEEbiography}[{\includegraphics[width=1in,height=1.25in,clip,keepaspectratio]{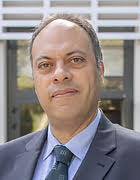}}]{Carl James Debono} (Senior Member, IEEE) received the B.Eng. degree (Hons.) in electrical engineering from the University of Malta, Malta, in 1997, and the Ph.D. degree in electronics and computer engineering from the University of Pavia, Italy, in 2000. From 1997 to 2001, he was a Research Engineer in the area of Integrated Circuit Design at the University of Malta. In 2001, he was appointed Lecturer with the Department of Communications and Computer Engineering, University of Malta, where he is currently a Professor. He currently serves as the Dean of the Faculty of Information and Communication Technology, University of Malta. Prof. Debono has participated in a number of local and European research projects in the area of communication systems and image/video processing. His research interests include multiview video coding, resilient multimedia transmission, and computer vision.
\end{IEEEbiography}

\begin{IEEEbiography}[\rotatebox{-90}{{\includegraphics[width=1in,height=1.25in,clip,keepaspectratio]{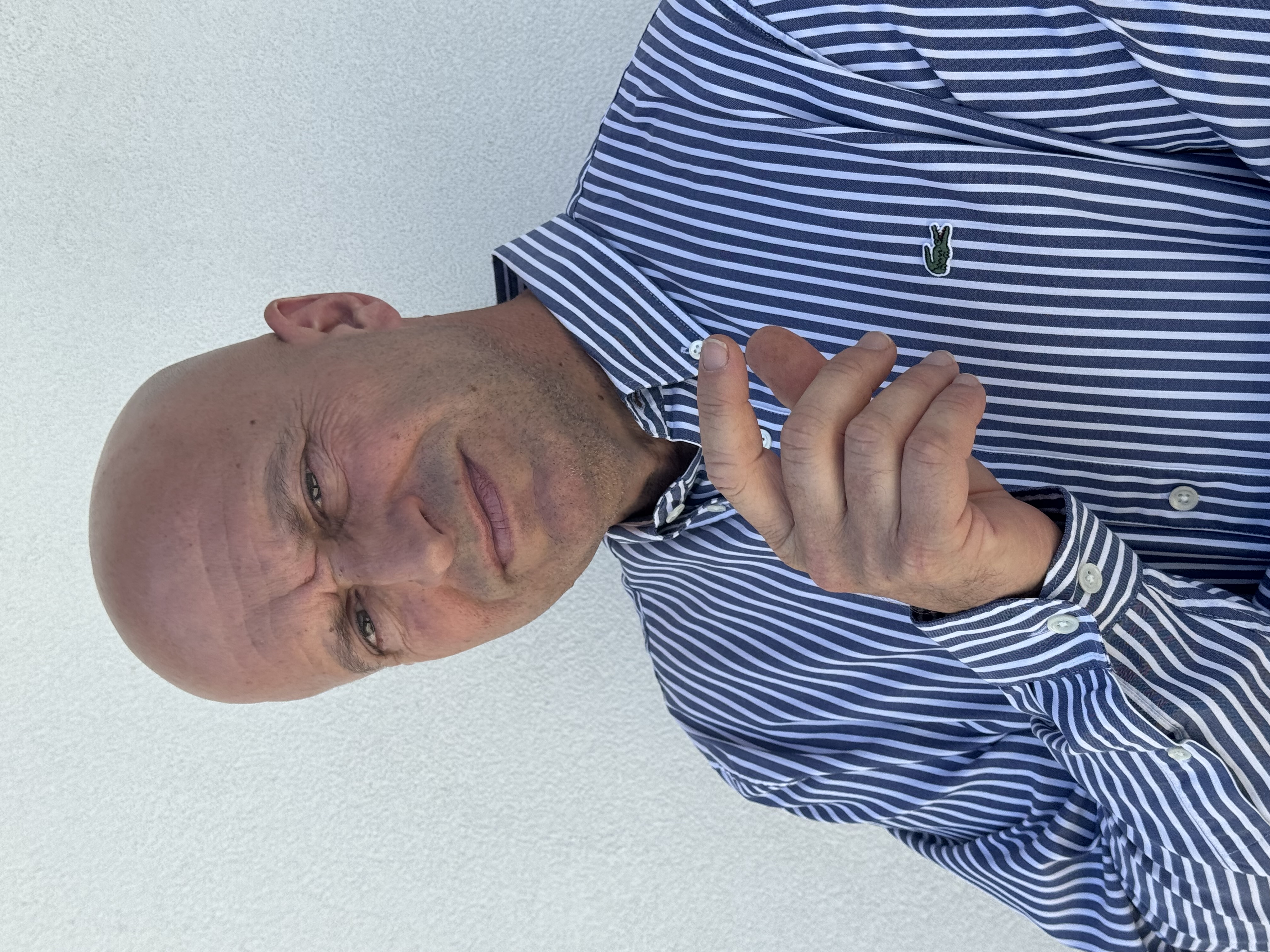}}}]{Saviour Formosa} is a senior academic at the University of Malta and consultant to national and international public-sector entities. Formerly Head of the Department of Criminology (2016–2020), he holds a PhD in spatio-temporal environmental criminology, an MSc in GIS, and a BA(Hons) in Sociology. His research focuses on spatio-temporal analysis, 3D scene reconstruction, advanced scanning technologies, and safety and security. He has acquired and led major EU and national projects including InMotion, SIntegraM, and SpatialTRAIN, securing over €80 million in funding. He directs immersive and digital-twin research through the Immersion Lab at the University of Malta.
\end{IEEEbiography}

\begin{IEEEbiography}[{\includegraphics[width=1in,height=1.25in,clip,keepaspectratio]{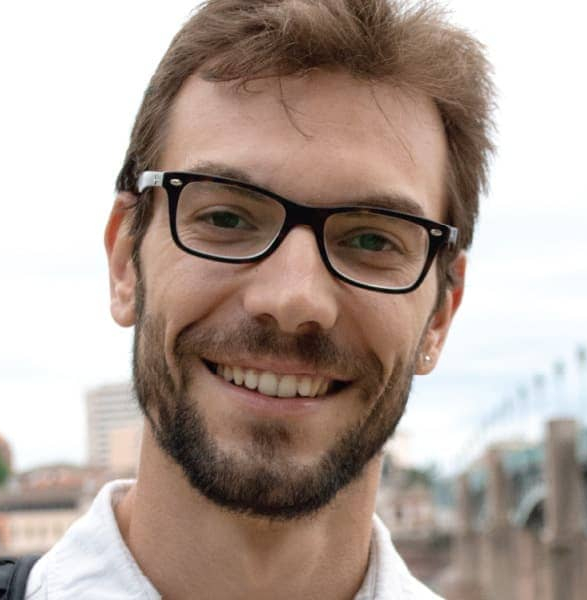}}]{Konstantinos Makantasis} (Member, IEEE)
received the Diploma, M.Sc., and Ph.D. degrees in computer engineering from the Technical University of Crete. He is a Lecturer with the Department of AI, University of Malta. He received the prestigious MSCA IF Widening Fellowship, to work on tensor-based machine learning methods for affect modeling. He has more than 80 publications in international journals and conferences on computer vision, signal and image processing, and machine learning. He is mostly involved and interested in computer vision, machine learning/pattern recognition, and probabilistic programming.
\end{IEEEbiography}

\end{document}